\documentclass[conference,10pt,letterpaper,twoside]{style/ieeeconf}

\IEEEoverridecommandlockouts    
\makeatletter 

\let\proof\@undefined
\let\endproof\@undefined

\let\NAT@parse\undefined
\makeatother

\usepackage[numbers, sort, compress]{natbib}
\usepackage{graphicx}
\usepackage{amsmath,amssymb}
\usepackage{caption}
\usepackage{tabularx}
\usepackage{subcaption}
\usepackage{soul}
\usepackage{hyperref} 
\usepackage{url} 

\usepackage{amsthm}
\usepackage{xcolor}

\usepackage{style/perl_acronyms}
\usepackage{style/perl_math}
\usepackage{style/perl_SIunits}
\usepackage{style/perl_misc}

\usepackage{fancyhdr}
\fancypagestyle{withfooter}{
  
  \fancyhead[L]{}
  \fancyhead[R]{}
  \fancyfoot[C]{\footnotesize Presented at the 2025 IEEE ICRA Workshop on Field Robotics}
}

\begin{document}

\pagestyle{withfooter}

\title{Towards Terrain-Aware Task-Driven 3D Scene Graph Generation \\ in Outdoor Environments}

\author{Chad~R.~Samuelson, Timothy~W.~McLain,~and Joshua~G.~Mangelson%
  \thanks{This work was partially funded under Office of Naval Research award number N00014-24-1-2301 and by the Center for Autonomous Air Mobility and Sensing (CAAMS), a National Science Foundation Industry-University Cooperative Research Center (IUCRC) under NSF award number 2139551, along with significant contributions from CAAMS industry members.}
  \noindent
  \thanks{C.~Samuelson, T.~McLain, and J.~Mangelson are at Brigham Young University. They can be reached at \texttt{\{chadrs2, mclain, mangelson\}@byu.edu}.}  %
}

\maketitle
\begin{abstract}

High-level autonomous operations depend on a robot's ability to construct a sufficiently expressive model of its environment. Traditional three-dimensional (3D) scene representations, such as point clouds and occupancy grids, provide detailed geometric information but lack the structured, semantic organization needed for high-level reasoning. 3D scene graphs (3DSGs) address this limitation by integrating geometric, topological, and semantic relationships into a multi-level graph-based representation. By capturing hierarchical abstractions of objects and spatial layouts, 3DSGs enable robots to reason about environments in a structured manner, improving context-aware decision-making and adaptive planning. Although most recent work has focused on indoor 3DSGs, this paper investigates their construction and utility in outdoor environments. We present a method for generating a task-agnostic metric-semantic point cloud for large outdoor settings and propose modifications to existing indoor 3DSG generation techniques for outdoor applicability. Our preliminary qualitative results demonstrate the feasibility of outdoor 3DSGs and highlight their potential for future deployment in real-world field robotic applications. 

\end{abstract}

\acresetall

\IEEEpeerreviewmaketitle

\section{Introduction}


As autonomous robots become increasingly adept at navigating and interacting with complex environments, the demand for efficient and effective structured world representations has grown. Over the years, numerous methods have been developed to enable accurate and robust real-time geometric localization and mapping~\cite{slam_murORB2,slam_isam25979641,slam_vinsmono8421746,slam_pairwiseconsist8460217,slam_liosam2020shan}. While traditional three-dimensional (3D) maps and point clouds provide rich geometric detail, they often lack the semantic and hierarchical organization required for high-level reasoning and context-aware planning.

Recent advancements in lightweight open-set language and image classification models have accelerated the development of truly autonomous robotic systems capable of mapping, reasoning, and planning in a manner more aligned with human reasoning. Large language models (LLMs) and vision-language models (VLMs) have played a key role in this progress, offering a unified framework that enhances modularity and adaptability across diverse environments and tasks~\cite{guConceptGraphsOpenVocabulary2024,werby23hovsg,qiu-song-peng-2024-wildlma,Devarakonda2024OrionNavOP,steinkeCollaborativeDynamic2025b}. Integrating semantic understanding from these models into geometric maps holds the potential to bridge the gap between structure and reasoning, enabling more intelligent and context-aware robotic behavior.

To this end, 3D scene graphs (3DSGs) have emerged as a powerful technique that enables hierarchical integration of both geometric and semantic information from commonly used robotic sensors~\cite{armeni3DScene2019b,rosinolKimera2020, rosinol3DDynamic2020,hughes2022hydra,bavle2022sgraphs+,bavle2025sgraphs20hierarchicalsemantic,guConceptGraphsOpenVocabulary2024,werby23hovsg,takmazSearch3DHierarchical2024,xu2024point2graphendtoendpointcloudbased,maggio2024Clio,Devarakonda2024OrionNavOP,qiu-song-peng-2024-wildlma,gaoEnhancingScene2024,xu2025tbhsu,kassab2024barenecessities,tangOpenINOpenVocabulary2025,2025topofield,bergUsingLanguage2022,greve2024curb,straderIndoorOutdoor2024,steinkeCollaborativeDynamic2025b}. The hierarchical structure of a 3DSG facilitates efficient context-aware task planning and navigation by offering a scalable and structured representation of the environment that enhances both perception and decision-making in autonomous systems.

\begin{figure}
   \centering
    \includegraphics[height=6cm]{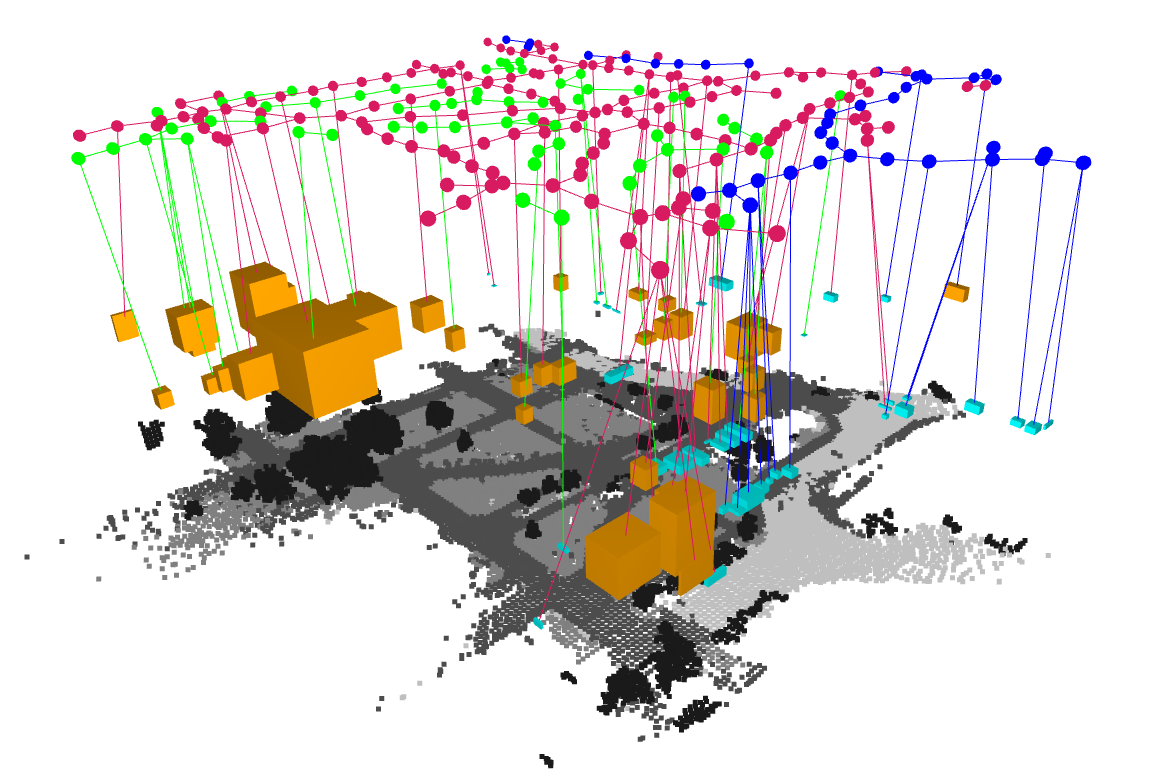}
\caption{Preliminary implementation of a task-driven terrain-aware semantic outdoor 3D scene graph. Level~1: A gray-scale metric-semantic point-cloud representation of the scene. Level~2: Object bounding-boxes for queried object types relevant to a specific task (tree [{\em orange}], car [{\em cyan}]). Level~3: Terrain-aware place nodes and topological connections for queried terrain types (sidewalk [{\em magenta}], grass [{\em green}], asphalt [{\em blue}]). }
    \vspace{-0.5cm}
    \label{fig:full_3dsg}
\end{figure}

Most existing approaches primarily focus on indoor 3DSG generation. While some efforts have been made to extend these methods to outdoor environments, the limited number of proposed techniques are typically application-specific, require prior scene knowledge, or lack a truly hierarchical scene interpretation~\cite{bergUsingLanguage2022,greve2024curb,straderIndoorOutdoor2024,rayTaskMotion2024,steinkeCollaborativeDynamic2025b}. Outdoor 3DSG generation presents additional unique challenges (compared to indoor 3DSG generation) due to each of the following reasons:
\begin{enumerate}
    \item The delineation of regions (clearly defined in indoor spaces, e.g. rooms/buildings) is ambiguous and task-dependent in outdoor scenes
    \item Terrain characterization (of minimal importance indoors) plays a fundamental role in outdoor navigation operations
    \item The level of dynamics of physical scene structure (as opposed to objects) in outdoor scenes is dramatically higher than in indoor environments
    \item Operational conditions are generally more significant due to changes in lighting, weather, and season
\end{enumerate} 
We seek to address these challenges in this and future work.

To effectively address the first challenge, a task-driven 3DSG generation approach is necessary. Task-driven 3DSGs hierarchically organize relevant information, enhancing robustness and efficiency for autonomous robotic applications in complex environments. Clio~\cite{maggio2024Clio} introduced a method for generating 3DSGs tailored to a given task or set of tasks. Building on Clio's work, we aim to extend this approach to outdoor environments by first constructing a task-agnostic metric-semantic map that can be repeatedly used based on specific tasks.

In this work, we present initial results towards building a terrain-aware outdoor 3DSG focusing on the first two challenges above (see Fig. \ref{fig:full_3dsg}). This paper describes the following contributions of our research:
\begin{itemize}
    \item An approach for task-agnostic open-set metric-semantic mapping built by fusing 3D LiDAR and camera sensor data;
    \item A novel terrain-aware semantic segmentation strategy for place node construction in outdoor 3DSGs using a general Voronoi diagram (GVD). 
\end{itemize}

\section{Related Work}



\subsection{Initial Indoor 3D Scene Graph Pipelines}

Scene graphs were initially developed to characterize the relationship between different objects in a given image~\cite{krishna2DSG2017}. Armeni et al.~\cite{armeni3DScene2019b} extended this concept to 3DSGs that graphically represent the spatial relationship between objects in a 3D scene. Their approach takes as input a 3D mesh built from RGB panoramic images. Semantic detectors are then used to detect indoor objects using multi-view consistency for robust 3D mesh classification. A hierarchical representation of the world is then built in the form of a four-layer graph, with layer 4 being the camera views, layer 3 the detected objects in the rooms, layer 2 the segmented rooms~\cite{armeni3DSemantic2016}, and layer 1 the root node. 

Subsequent approaches refined and expanded this framework.~\citet{rosinolKimera2020} introduced Kimera, a full 3DSG pipeline that takes raw RGB-D images and IMU as input and utilizes modern robotic SLAM methods and object classifiers to construct a dense metric-semantic mesh. Later enhancements added the ability to track dynamic agents as they move throughout the scene~\cite{rosinol3DDynamic2020}. \citet{bavle2022sgraphs+,bavle2025sgraphs20hierarchicalsemantic} proposed situational graphs (S-Graphs) that directly fuse modern factor-graph-based SLAM methods and LiDAR-based scene-graph generation, but lacked object classification capabilities. \citet{hughes2022hydra} later introduced Hydra, which builds upon  Kimera while focusing on real-time 3DSG generation. 

\subsection{Recent Fusion of Foundation Models with Scene Graphs}

The recent explosion of LLMs and VLMs has led to multiple techniques that seek to leverage these tools to enhance both 3DSG generation and utilization~\cite{guConceptGraphsOpenVocabulary2024,werby23hovsg,takmazSearch3DHierarchical2024,maggio2024Clio,xu2024point2graphendtoendpointcloudbased,Devarakonda2024OrionNavOP,qiu-song-peng-2024-wildlma,gaoEnhancingScene2024,xu2025tbhsu,kassab2024barenecessities,tangOpenINOpenVocabulary2025}. 

LLMs are deep learning models that have been trained on vast amounts of text data to understand and generate human-like language. Beyond human-like text generation, LLMs have demonstrated abilities to interpret, understand, and relate abstract language concepts and ideas~\cite{guConceptGraphsOpenVocabulary2024}.

Many techniques~\cite{guConceptGraphsOpenVocabulary2024,Devarakonda2024OrionNavOP,qiu-song-peng-2024-wildlma,tangOpenINOpenVocabulary2025} seek to utilize LLMs such as GPT-4o~\cite{OpenAI2024GPT4o} and LLaMa~\cite{touvron2023llamaopenefficientfoundation} in 3DSG frameworks to infer relationships between objects. \citet{Devarakonda2024OrionNavOP} uses GPT4 and LLaMa LLMs to perform task planning within their indoor 3DSG. While LLMs enable high-level reasoning in 3DSGs, their computational demands and high costs limit real-time use on robotic platforms, restricting them to cloud or offline processing.

VLMs are trained using combinations of image and textual data and in many cases enable mapping between these essential modalities. Unlike LLMs, VLMs are more lightweight, allowing for real-time execution. The encoder portion of the popular open-source CLIP~\cite{radfordLearningTransferable2021} model provides the ability to embed both image and language into a shared latent space. The internet-scale training data used to create these models enable open-set (or wide-set) object classification and query at scales not previously possible. Moreover, while CLIP was originally trained on full images with associated captions, recent studies have demonstrated its surprising ability to classify masked objects within images~\cite{jatavallabhula2023Conceptfusion,Devarakonda2024OrionNavOP,guConceptGraphsOpenVocabulary2024,kassab2024barenecessities,maggio2024Clio,takmazSearch3DHierarchical2024,werby23hovsg,xu2024point2graphendtoendpointcloudbased,tangOpenINOpenVocabulary2025}.

These methods typically use class-agnostic segmentation masks from foundational segmentation models, such as the Segment Anything Model (SAM) and its variants~\cite{kirillov2023sam,ravi2024sam2,zhao2023fastsam,zhang2023mobilesam}, to extract object masks. Rather than using object masks, Yolo-World~\cite{cheng2024YOLOWorld} integrates the CLIP text encoder directly into the efficient YOLO framework, thus allowing the extraction of image bounding boxes corresponding to arbitrary open-set text queries.

Most VLM-based 3DSG methods use VLMs to generate semantic embeddings for objects and regions during scene reconstruction, then construct a 3DSG based on these embeddings and the geometric relationships between objects and regions. However,~\citet{xu2025tbhsu} integrates scene reconstruction, semantic embedding, and 3DSG generation into a unified, end-to-end transformer-based framework. While most approaches focus on embedding entire objects, Search3D~\cite{takmazSearch3DHierarchical2024} incorporates part segmentation, embedding object parts within the 3DSG. Meanwhile, \citet{xu2024point2graphendtoendpointcloudbased} employs VLMs exclusively for room classification, relying on separate geometric and semantic models for object segmentation and classification. For an overview and evaluation of common VLM techniques in 3DSGs, we refer the reader to~\cite{kassab2024barenecessities}.

Of most relevance to our work, \citet{maggio2024Clio} introduced Clio which uses the CLIP language and image encoder~\cite{radfordLearningTransferable2021} and a pre-specified task to generate a 3DSG that is task-driven, tracking only the objects and information relevant to the specified task. In this work,~\citet{maggio2024Clio} conclude that the optimal 3DSG structure is dependent on the task to be carried out. This insight is of particular relevance as we work to develop 3DSG techniques for modeling of unstructured outdoor scenes.

\subsection{Outdoor 3D Scene Graphs}

Significantly less research has focused on the generation of 3DSGs for outdoor environments. This is primarily due to the unstructured nature and variability of outdoor environments. 

Initially, hierarchical semantic representations of outdoor environments remained application-specific. In aerial settings, \citet{bergUsingLanguage2022} designed a semantic hierarchy map generated from the open-source satellite image dataset, OpenStreetMap (OSM), to perform task planning. For urban applications, \citet{greve2024curb} introduced a multi-agent centralized 3DSG framework that relies solely on LiDAR data for both map construction and closed-set semantic embeddings. Their urban 3DSG includes application-specific layers such as lane graphs, vehicles, intersections, and roads. \citet{steinkeCollaborativeDynamic2025b} has extended this previous method to include camera data for open-set 3DSG mapping.

Generalizable outdoor 3DSG generation remains an open problem. To this end, \citet{straderIndoorOutdoor2024} used an LLM-driven ontology to define general object and region relationships and classification for indoor or outdoor environments. This approach generalizes well when key items help uniquely identify concepts. For instance, detecting a toilet strongly indicates a bathroom region. However, it struggles with more abstract distinctions, such as differentiating a family room from a living room or a park from a backyard. Given these limitations in seeking to build a more generalizable outdoor 3DSG, our work adopts a task-dependent 3DSG approach like~\cite{maggio2024Clio}.


\subsection{Segmentation and Classification of Terrain}

While the open-set capabilities of VLMs are widely-touted, they still regularly fail to classify out-of-distribution classes that are not sufficiently represented within the training distribution. Moreover, we have observed that  VLMs typically struggle with terrain classification, potentially due to the fact that the captions they were trained on primarily focused on describing objects within a scene rather than background.

To enhance outdoor semantic understanding, which VLMs lack, training a new classification model requires outdoor terrain data. However, few datasets generalize well across various outdoor robotic settings. Early terrain-based datasets focused on trails and off-road driving but lack broader applicability~\cite{valadaFreiburgForest2017a,maturanaRealTimeSemantic2018a,wignessRUGDDataset2019,jiangRELLIS3DDataset2021}. More recent datasets, such as TAS500~\cite{metzger2021}, GOOSE~\cite{mortimer2024goose}, GOOSE-Ex~\cite{hagmanns2024gooseEx}, and WildScenes~\cite{vidanapathiranaWildScenesbenchmark2024}, still face challenges in handling common outdoor scenes, indicating the need for a truly general terrain dataset.

\section{Terrain-Aware Task-Driven Outdoor 3D Scene Graph Generation}

\begin{figure*}[t]
   \centering
    \includegraphics[width=\textwidth]{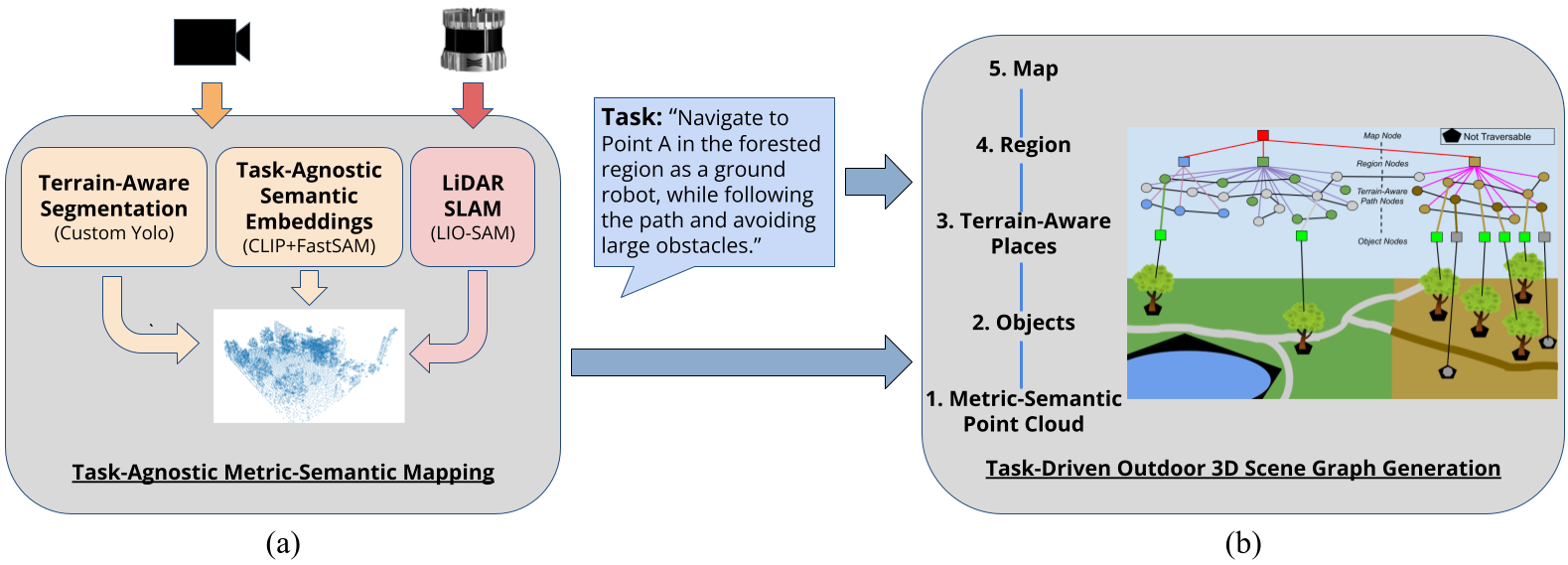}
    \caption{(a) \textit{Task-Agnostic Metric-Semantic Mapping.} Mapping steps: (1) Extract terrain and class-agnostic semantics from RGB imagery. (2) Generate a global 3D point-cloud map using LIO-SAM. (3) Fuse these results to produce a CLIP-based semantic embedding for each sparse point in the 3D point-cloud map. (b) \textit{Task-Driven 3D Scene Graph Generation.} Upon receipt of a task specification, a unique 3DSG structured specifically to support the task at hand is built.}
    \vspace{-0.5cm}
    \label{fig:3dsg_overview}
\end{figure*}

In this paper, we present preliminary results towards terrain-aware task-driven outdoor scene graph generation. We first present a high-level overview of our full framework (as shown in Fig. \ref{fig:3dsg_overview}) and then provide details on  the initial stages of our method in the following sections.

Building on the insights introduced by~\citet{maggio2024Clio}, we believe that 3DSGs are best structured around the requirements of a specific task. However task-driven 3DSGs are limited in their ability to generalize as task requirements and goals change. Ideally, a single map of a scene should be applicable to any task, even if the task is unknown before mapping takes place.

As such, we propose to break the scene graph generation process into the following two phases: (1) task-agnostic metric-semantic mapping, and (2) task-driven 3DSG generation. This separation enables reuse of the same metric-semantic map across multiple tasks, while allowing generation of task-driven 3DSGs for each task. While these phases can be performed iteratively in real time, our current focus is on completing the first phase before advancing to the second.

\subsection{Phase 1: Task-Agnostic Metric-Semantic Mapping}
\label{meth:task_agnostic_pc}

We first construct a general task-agnostic metric-semantic map for the scene that will form the backbone for future task-driven 3DSG generation. To accomplish this, we must develop a model of the environment that includes (1) an accurate metric understanding of the scene and (2) task and class agnostic semantic information. 

While the general principles in our framework above could be implemented using a variety of sensor configurations, in this work we assume our system is provided with time-synchronized and extrinsically calibrated RGB images, high-rate IMU measurements, and dense 3D LiDAR point-clouds. This input data is fused to create a global point-cloud representation of the environment that includes class-agnostic semantic embeddings for each point. 


\subsubsection{Metric Localization and Mapping} 
To perform metric localization and mapping, any accurate simultaneous localization and mapping (SLAM) technique that results in a global 3D point-cloud can be used. In our initial tests, we utilize LIO-SAM~\cite{slam_liosam2020shan}, a factor-graph based SLAM system that builds upon LOAM \cite{slam_loam2014zhang} and incorporates both inertial and loop-closure measurements into a global pose graph. This pose-graph is then optimized using non-linear least-squares techniques, resulting in a maximum likelihood trajectory of the robot and a sparsified global point-cloud map of the scene.

\subsubsection{Semantic Feature Extraction}
To obtain semantic information, we simultaneously process overlapping RGB imagery of the scene to obtain latent feature vectors that encode the semantic content of each segment. To reliably classify terrain, we use a fine-tuned YOLOv11 network~\cite{yolov8_ultralytics} that segments and classifies common terrain types. We then save these segments and mask out the terrain regions from the RGB images. FastSAM~\cite{zhao2023fastsam} is then used to generate class-agnostic masks for the remaining portion of the image. A latent feature vector for each FastSAM mask is then obtained using CLIP~\cite{radfordLearningTransferable2021}. Each terrain text label from the YOLO model is also encoded using CLIP's text encoder to obtain corresponding semantic embeddings. 

\subsubsection{LiDAR Point and Semantic Feature Association}
To associate LiDAR points with latent semantic feature vectors, we back-project the time-synchronized LiDAR scan into the RGB image frame using precomputed intrinsic and extrinsic calibrations. LiDAR points are then associated to the embeddings corresponding to the image segments in which they appear. KD-Trees~\cite{bentley1975kdtrees} are then employed to match sparse global map points with dense LiDAR scan points, allowing us to store semantic embeddings for each sparse global map point. As new time-synchronized scans and images are acquired, CLIP embeddings are computed iteratively and averaged per point to enhance semantic robustness through multi-view aggregation.  Future work will explore more efficient methods for tracking global map semantics rather than point-by-point. 

Once created, the completed task-agnostic metric-semantic point-cloud can be used to support the generation of various 3DSGs depending on the task at hand.

\subsection{Phase 2: Task-Driven 3D Scene Graph Generation}

Given a specific task, we now seek to build a task-dependent hierarchical 3DSG based upon the metric-semantic point-cloud map developed in Sec.~\ref{meth:task_agnostic_pc}. Our proposed graph structure is divided into five discrete levels outlined as follows: 1) metric-semantic point cloud built from phase 1, 2) object nodes, 3) terrain-aware place nodes, 4) region nodes, 5) map (root) nodes.

\subsubsection{Metric-Semantic Point Cloud}

The base layer (layer 1) contains a metric-semantic point cloud of the scene. We generate this layer by extracting all terrain and task-related object point cloud points from the metric-semantic point cloud built in phase 1. Sec.~\ref{meth:objectnodes} and \ref{meth:terrain_placenodes} explain the prompting process used on the point cloud to extract relevant points.

\subsubsection{Object Nodes}
\label{meth:objectnodes}

The second layer of the scene graph encodes the bounding boxes of the task-relevant objects. This is done by computing the cosine similarity between the CLIP-embeddings of each task-relevant object from the task prompt with the embeddings of each point in the metric-semantic point cloud. If the cosine similarity between the object and the point are above a given threshold, $\alpha$, then we assign that point to the associated class. Finally, we cluster points that are geometrically close together and semantically consistent using DBSCAN to get object bounding boxes. Each bounding box is added to the 3DSG as a semantic object node. 

\subsubsection{Terrain-Aware Place Nodes}
\label{meth:terrain_placenodes}

\begin{figure}[t!]
    \centering
    \begin{subfigure}[b]{0.48\columnwidth}
         \centering
         \includegraphics[width=\columnwidth]{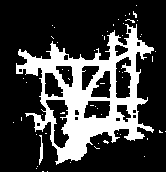}
         \captionsetup{skip=5pt}
         \caption{}
         \label{fig:occ_grid}
    \end{subfigure}
    \begin{subfigure}[b]{0.48\columnwidth}
         \centering
         \includegraphics[width=\columnwidth]{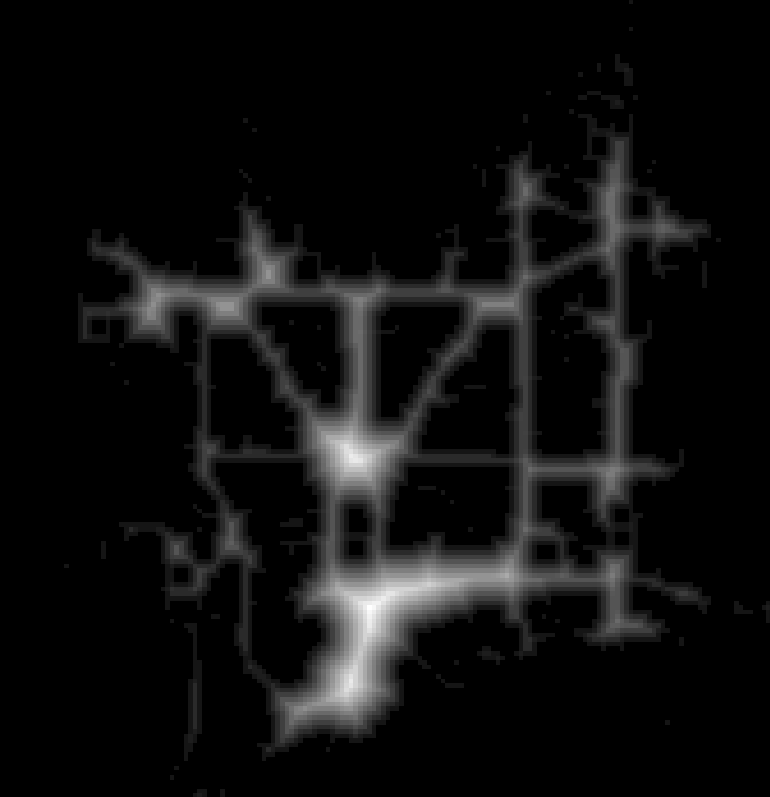}
         \caption{}
         \label{fig:dist_map}
    \end{subfigure}
    \begin{subfigure}[b]{0.48\columnwidth}
         \centering
         \includegraphics[width=\columnwidth]{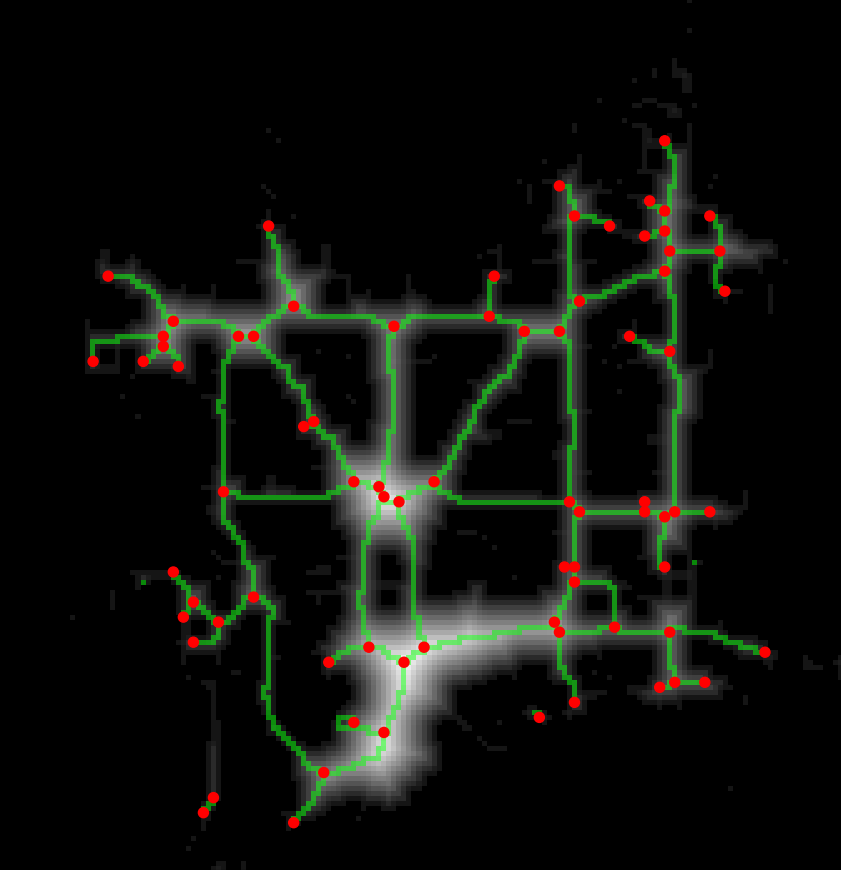}
         \caption{}
         \label{fig:init_nodes}
    \end{subfigure}
    \begin{subfigure}[b]{0.48\columnwidth}
         \centering
         \includegraphics[width=\columnwidth]{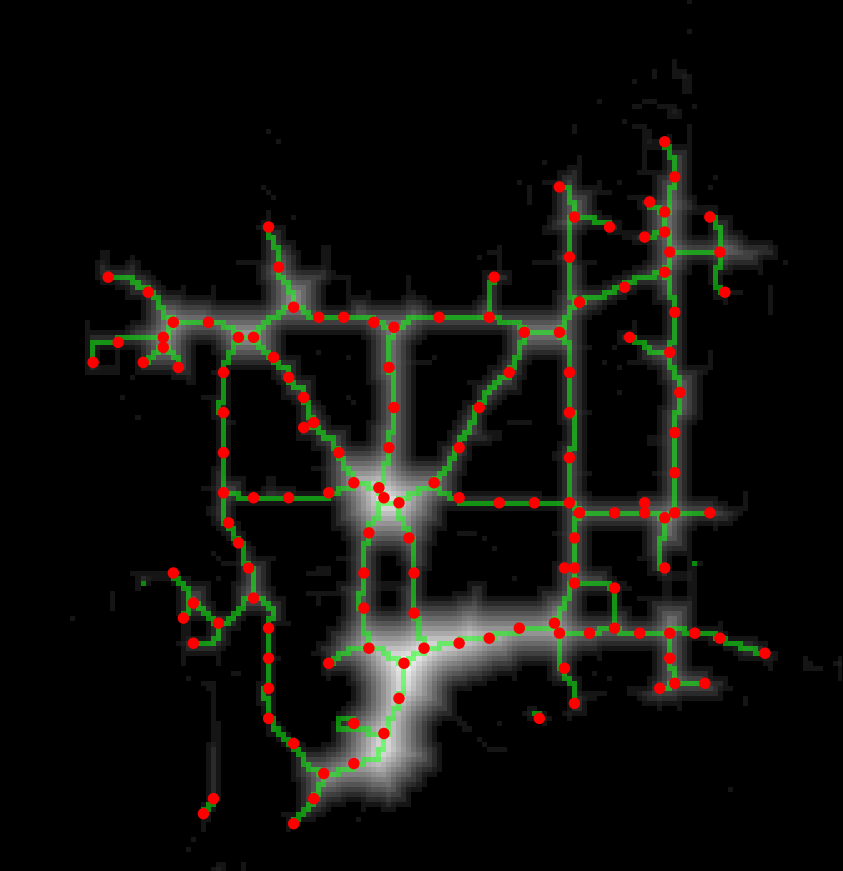}
         \caption{}
         \label{fig:finished_nodes}
    \end{subfigure}
    \caption{GVD process: (a) Initial occupancy grid of sidewalk terrain; (b) Distance map - whiter pixels represent larger distances; (c) Pruned GVD with initial nodes [{\em red}] and edges [{\em green}]; (d) Final GVD with nodes [{\em red}] and edges [{\em green}].}
    \vspace{-0.5cm}
    \label{fig:gvd_process}
\end{figure}

The third layer encodes the terrain into a generalized Voronoi diagram (GVD) by terrain type where the GVD nodes and edges are associated with the 3DSG place nodes, similar to \cite{hughes2022hydra,rayTaskMotion2024}. For each terrain type, we generate a binary two-dimensional (2D) grid map of the scene where a value of one encodes presence of the specified terrain type at that point and zero represents absence of the terrain at that point. We then apply basic dilation and erosion techniques to smooth out noisy misidentified cells (see Fig. \ref{fig:occ_grid}). Simultaneously, we compute the GVD alongside a distance map using the brushfire algorithm based on~\cite{lauEfficientgridbased2013} (with the distance map shown in Fig. \ref{fig:dist_map}). The GVD consists of 2D cells, each classified as either a node or an edge. Initially, all cells are edges. After pruning the GVD~\cite{lauEfficientgridbased2013}, we designate certain cells as nodes--specifically those at junctions and corners--while ensuring that no nodes are orthogonally adjacent (see Fig. \ref{fig:init_nodes}). Because our GVDs are 2D, we define a {\em junction} as a cell with three or more orthogonally or diagonally adjacent cells, while a {\em corner} is a cell with exactly one orthogonally adjacent cell. Next, we iteratively refine the GVD through two main steps: (1) a flood-fill algorithm assigns each edge to its closest node, and (2) we split edges between connected nodes by inserting new nodes based on the following criteria. If an edge in the chain between two nodes deviates from the straight line connecting them by more than a given Euclidean distance threshold, and it is the most deviated edge in the chain, it is converted into a node. Also if the distance between two connected nodes exceeds a specified threshold, an edge at the midpoint is converted into a node. This iterative process continues until no new nodes are added or until a predefined number of iterations is reached, following~\cite{hughes2022hydra} (see Fig. \ref{fig:finished_nodes}). The final GVD graph forms the terrain-aware place nodes for that terrain type. This process is repeated for each terrain type. Object nodes from layer 2 are connected to their spatially closest terrain node.

\subsubsection{Region Nodes}
The fourth layer encodes the task-driven region nodes by grouping together related terrain and objects. We plan to group terrain and objects together into regions using the AIB algorithm presented in \citet{maggio2024Clio}. We expect promising region clustering results that can vary vastly depending on the task by using this information-theoretical approach. This remains as future work to fully implement this layer of the outdoor 3DSG. 

\subsubsection{Map Node}

The fifth (and highest) layer groups all the task-driven region nodes together within the environment.

\section{Experiments}

\begin{figure}[t!]
    \centering
    \begin{subfigure}[b]{0.525\columnwidth}
         \centering
         \includegraphics[width=\columnwidth]{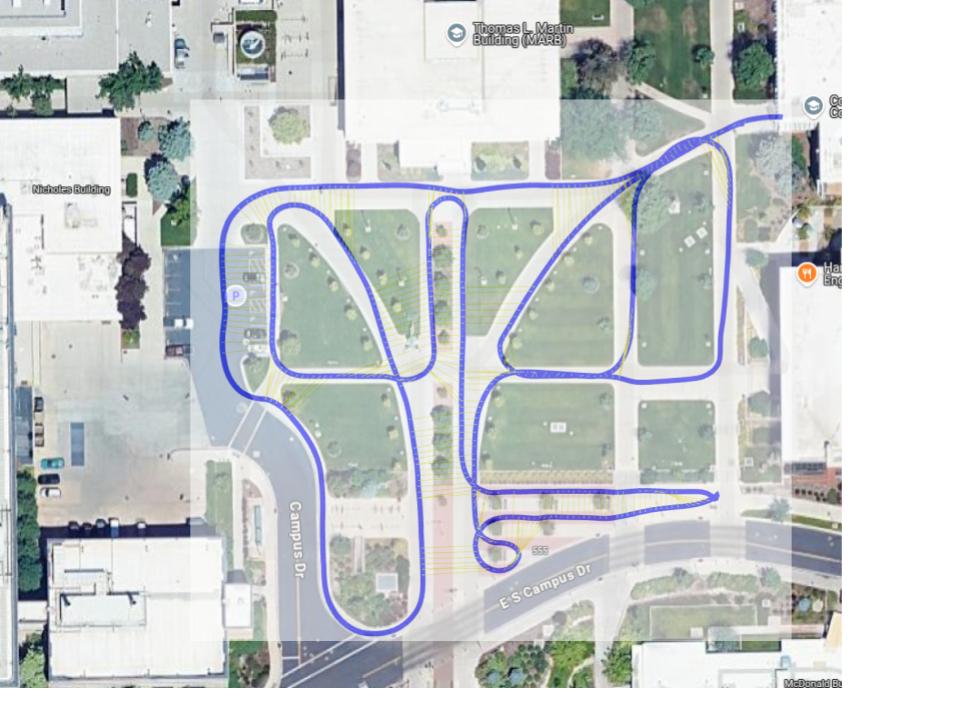}
         \caption{}
         \label{fig:liosam_path}
    \end{subfigure}
    \begin{subfigure}[b]{0.455\columnwidth}
         \centering
         \includegraphics[width=\columnwidth]{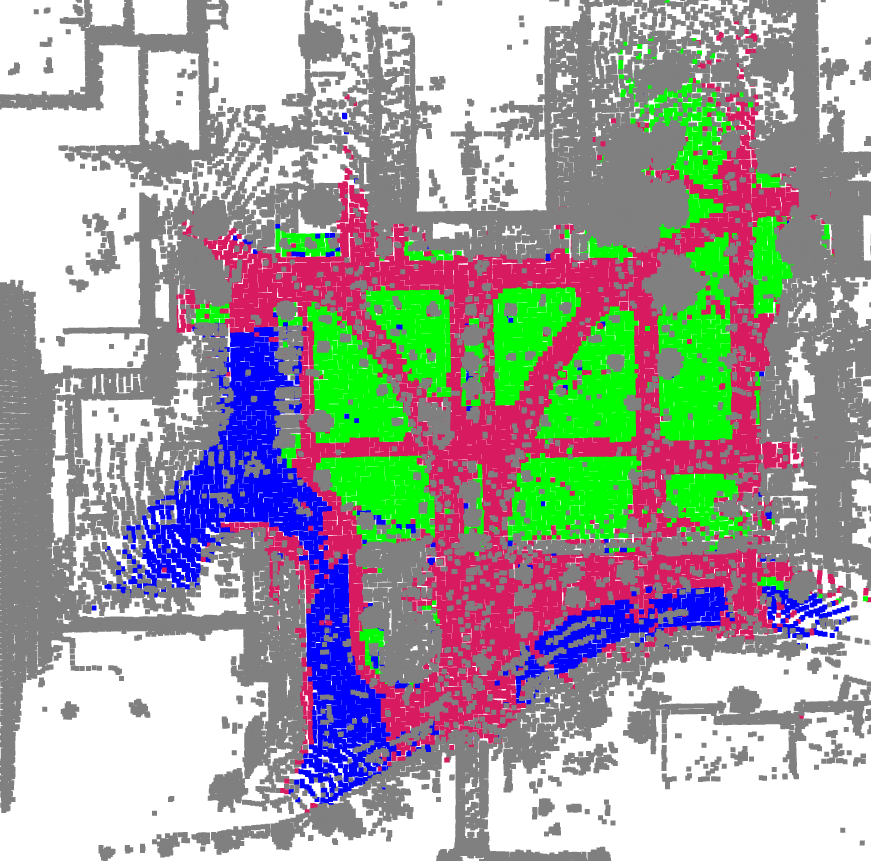}
         \caption{}
         \label{fig:terrain_pc_map}
    \end{subfigure}
    \caption{Terrain point-cloud results: (a) LIO-SAM path overlayed on satellite image; (b) terrain classified point cloud (sidewalk [{\em magenta}], grass [{\em green}], asphalt [{\em blue}]).}
    \vspace{-0.3cm}
    \label{fig:semantic_pc_tests}
\end{figure}

For our experiments, we used data collected from an OAK-D LR camera, a 128-beam OS1 Ouster LiDAR, and the Ouster-embedded IMU. To test the initial stages of our framework, we focused on a simpler outdoor environment with varying terrain types: a section of the southeast part of BYU campus. Fig. \ref{fig:liosam_path} shows the path traveled for this experiment extracted from LIO-SAM and overlayed on a satellite image view of the test area. Future work will test this framework in more complex outdoor environments.

\subsection{Terrain Classification}

To qualitatively evaluate the terrain classification results, we prompted the metric-semantic point cloud with only the trained terrain classes: sidewalk [{\em magenta}], grass [{\em green}], and asphalt [{\em blue}]. The results of these terrain prompts are shown in Fig. \ref{fig:terrain_pc_map}. Some gray points (i.e., unclassified points) remain in the classified regions, likely due to dynamic objects in the scene (e.g., walking pedestrians and moving cars) and the point-by-point classification approach, which lacks the spatial coherence of a clustering-based approach in semantic classification. Additionally, some incorrectly classified terrain regions can be improved on by training the YOLOv11 model on more terrain data.


\subsection{Open-Set Object Prompting}
\label{exp:openset_obj_prompting}

\begin{figure}
   \centering
    \includegraphics[height=4.5cm]{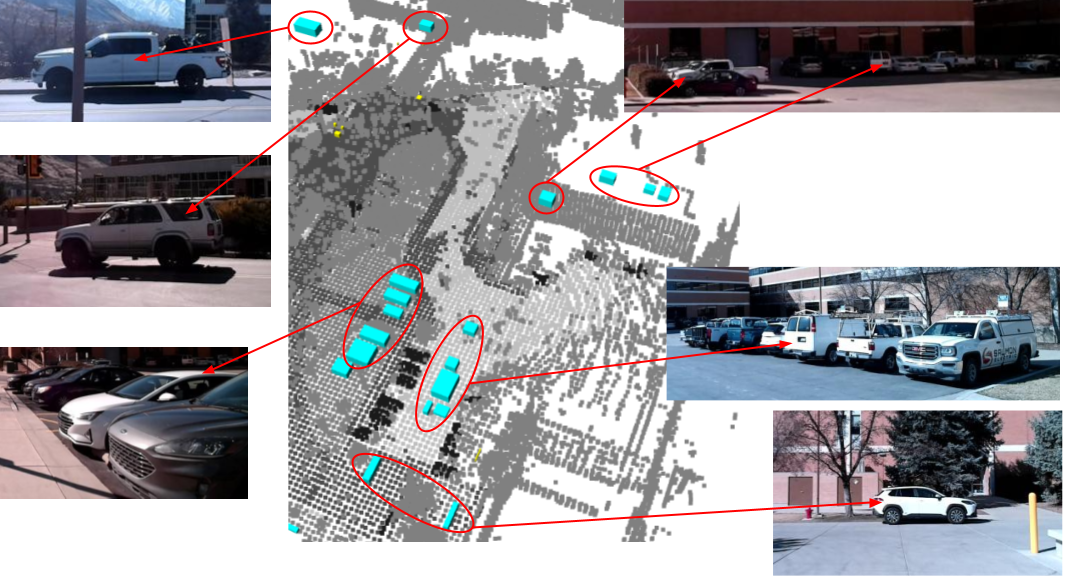}
    \caption{Matching images associated with the ``image of a car" object prompt. Car detections shown by black point-cloud points and a hovering cyan bounding box.}
    \vspace{-0.5cm}
    \label{fig:prompt_car_img_match}
\end{figure}

To evaluate the promptable capabilities of our metric-semantic point cloud, we introduced the text queries ``image of a tree" (orange) and ``image of a car" (cyan). Fig. \ref{fig:prompt_car_img_match} shows a few detected car objects in the point cloud and associated images from the RGB camera. We see that our method correctly identified prompted car objects in the metric-semantic point cloud. A few failed cases are shown in the upper right image of Fig. \ref{fig:prompt_car_img_match}, where some dark-colored cars in the shade are visibly harder to detect, causing missed detections. An incorrect detection is also shown in the bottom right image of Fig. \ref{fig:prompt_car_img_match} where two separate bounding boxes relate to the same car. This is likely due to the fact that directly using a FastSAM mask and its CLIP embedding projected into the 3D point cloud results in the mask's bordering points to be cast on the scenery behind the masked object. We plan to improve these results in future work by applying DBSCAN to the masked points and applying the semantic embeddings only to the largest cluster of those points, thus removing the border-projected points.

\subsection{Terrain-Aware 3D Scene Graph}

\begin{figure}[t!]
    \centering
    \begin{subfigure}[b]{0.99\columnwidth}
         \centering
         \includegraphics[width=\columnwidth]{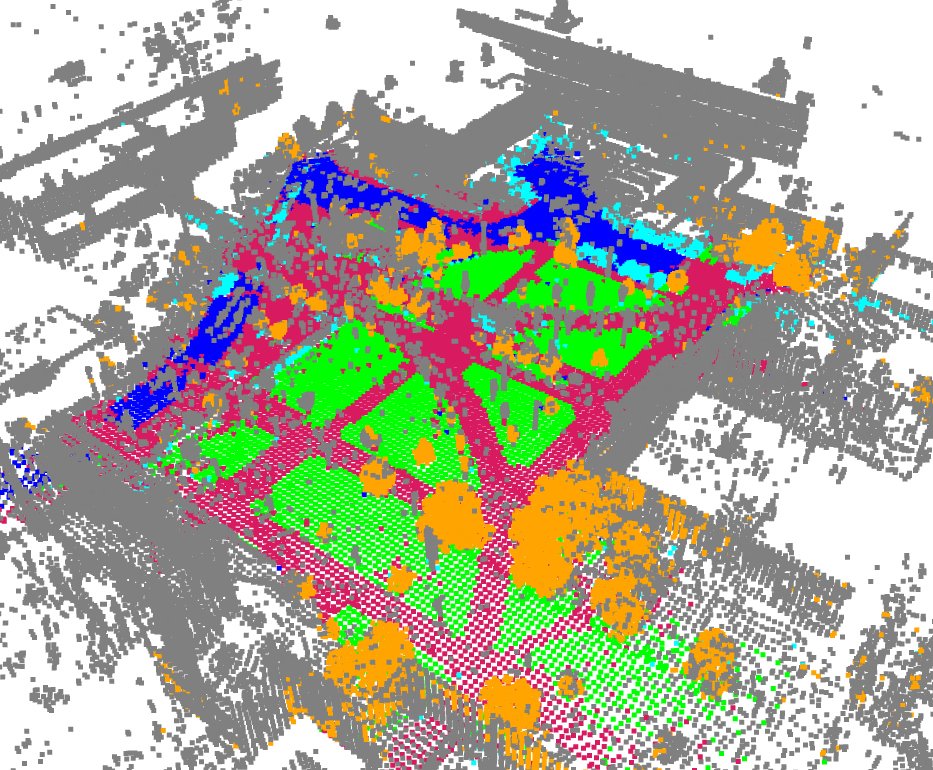}
         \captionsetup{skip=5pt}
         \caption{}
         \label{fig:prompted_pc}
    \end{subfigure}
    \begin{subfigure}[b]{0.99\columnwidth}
         \centering
         \includegraphics[width=\columnwidth]{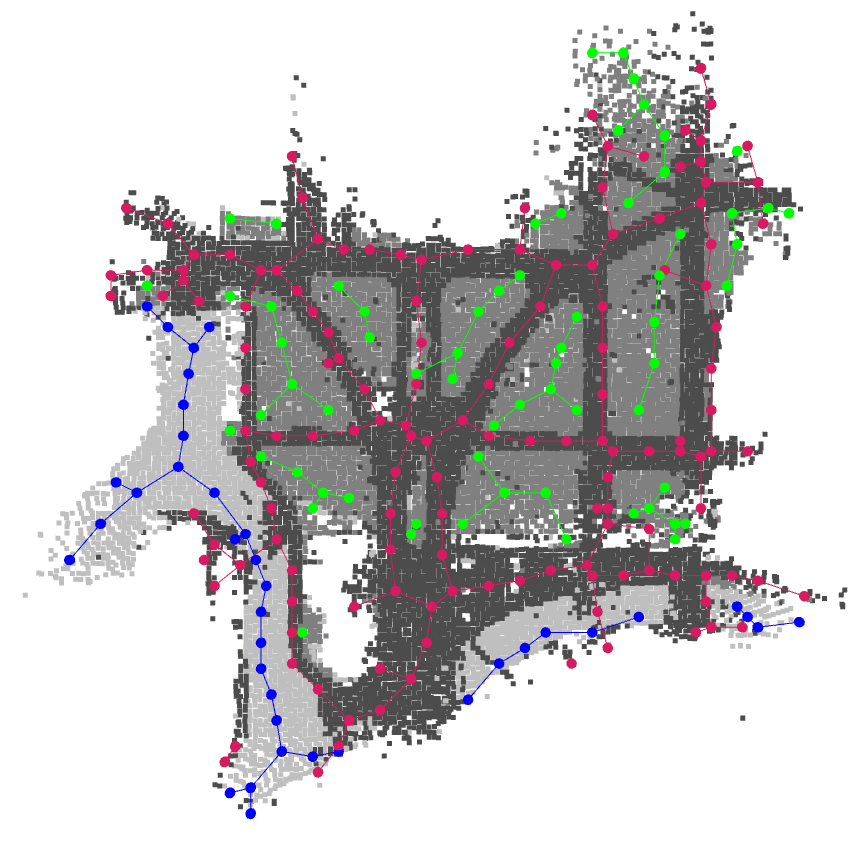}
         \caption{}
         \label{fig:terrain_gvd}
    \end{subfigure}
    \caption{Demonstration of the promptable nature of task-agnostic metric-semantic map and terrain-aware GVD place node layer: (a) Full metric-semantic point cloud colored by terrain and object prompts. Terrain: ``sidewalk" [{\em magenta}], ``grass" [{\em green}], ``asphalt" [{\em blue}]. Objects: ``image of a tree" [{\em orange}], ``image of a car" [{\em cyan}]. (b) GVD place node layer colored by terrain type (following coloring scheme from (a)).}
    \vspace{-0.5cm}
    \label{fig:demonstration_prompts_terrain_gvd}
\end{figure}

Our final experiment integrates all methods presented in Sec.~\ref{meth:objectnodes} -- \ref{meth:terrain_placenodes} into a single 3DSG. Fig. \ref{fig:prompted_pc} displays the original full global point cloud generated by LIO-SAM. Semantics are embedded into every point in the point cloud, as discussed in Sec.~\ref{meth:task_agnostic_pc}. 

To highlight the terrain types, we prompt the point cloud with the class names ``sidewalk," ``grass," and ``asphalt," coloring all matching points with a cosine similarity above $0.95$. This high threshold is chosen because the prompts exactly match the same CLIP text embeddings from the YOLOv11 model labels during the metric-semantic point cloud generation phase, but is below $1.0$ because of the effects of averaging other CLIP embeddings with these points. Next, we prompt the point cloud for the objects: ``image of a tree" and ``image of a car," applying a cosine similarity threshold of $\alpha=0.28$. The literature found higher cosine similarities when prompts included ``image of a \{\}"~\cite{xu2024point2graphendtoendpointcloudbased}. A lower threshold for text-to-image mapping is consistent with the literature~\cite{maggio2024Clio}. Qualitative results show that most trees were successfully detected, as indicated by the orange points in Fig. \ref{fig:prompted_pc}. However, some trees were missed, likely due to the challenges discussed in Sec.~\ref{exp:openset_obj_prompting}, as well as seasonal factors. Since the dataset was collected in February, small leafless trees look less tree-like.

Fig. \ref{fig:full_3dsg} shows the results of the semantically prompted point cloud, from Fig. \ref{fig:prompted_pc}, built into a terrain-aware hierarchical 3DSG. Only semantically classified points are retained in the 3DSG for layer 1 and are differentiated by different shades of gray: objects (black), sidewalk (dark gray), grass (normal gray), asphalt (light gray). Layer 2 contains the tree and car objects previously discussed. These object bounding box nodes are connected to the nearest GVD node in the place nodes layer. Layer 3 consists of the terrain-aware place nodes following the same coloring scheme as Fig. \ref{fig:prompted_pc} and \ref{fig:terrain_pc_map}. Fig. \ref{fig:terrain_gvd} is a top-down view of the GVD terrain-aware place nodes above the semantically associated 3D point cloud points. These results demonstrate the promising potential of an outdoor 3DSG method for semantically and geometrically understanding outdoor environments leading to the feasibility of truly autonomous robotic tasks in complex outdoor environments. 

\section{Conclusion}

In this paper we have shown progress towards the development of a terrain-aware outdoor 3DSG generation framework. We have shown promising results for semantically classifying outdoor objects and terrain. We have also demonstrated the development of the bottom three layers of our 3DSG. We see the development of outdoor scene graphs as a promising and exciting research direction and we plan to build upon the work presented here in future work.


\begin{thebibliography}{51}
\providecommand{\natexlab}[1]{#1}
\providecommand{\url}[1]{#1}
\csname url@samestyle\endcsname
\providecommand{\newblock}{\relax}
\providecommand{\bibinfo}[2]{#2}
\providecommand{\BIBentrySTDinterwordspacing}{\spaceskip=0pt\relax}
\providecommand{\BIBentryALTinterwordstretchfactor}{4}
\providecommand{\BIBentryALTinterwordspacing}{\spaceskip=\fontdimen2\font plus
\BIBentryALTinterwordstretchfactor\fontdimen3\font minus \fontdimen4\font\relax}
\providecommand{\BIBforeignlanguage}[2]{{%
\expandafter\ifx\csname l@#1\endcsname\relax
\typeout{** WARNING: IEEEtranN.bst: No hyphenation pattern has been}%
\typeout{** loaded for the language `#1'. Using the pattern for}%
\typeout{** the default language instead.}%
\else
\language=\csname l@#1\endcsname
\fi
#2}}
\providecommand{\BIBdecl}{\relax}
\BIBdecl

\bibitem[Mur-Artal and Tard\'os(2017)]{slam_murORB2}
R.~Mur-Artal and J.~D. Tard\'os, ``{ORB-SLAM2}: an open-source {SLAM} system for monocular, stereo and {RGB-D} cameras,'' \emph{IEEE Transactions on Robotics}, vol.~33, no.~5, pp. 1255--1262, 2017.

\bibitem[Kaess et~al.(2011)Kaess, Johannsson, Roberts, Ila, Leonard, and Dellaert]{slam_isam25979641}
M.~Kaess, H.~Johannsson, R.~Roberts, V.~Ila, J.~Leonard, and F.~Dellaert, ``{iSAM2}: Incremental smoothing and mapping with fluid relinearization and incremental variable reordering,'' in \emph{Proceedings of the IEEE International Conference on Robotics and Automation (ICRA)}, 2011, pp. 3281--3288.

\bibitem[Qin et~al.(2018)Qin, Li, and Shen]{slam_vinsmono8421746}
T.~Qin, P.~Li, and S.~Shen, ``{VINS-Mono}: A robust and versatile monocular visual-inertial state estimator,'' \emph{IEEE Transactions on Robotics}, vol.~34, no.~4, pp. 1004--1020, 2018.

\bibitem[Mangelson et~al.(2018)Mangelson, Dominic, Eustice, and Vasudevan]{slam_pairwiseconsist8460217}
J.~G. Mangelson, D.~Dominic, R.~M. Eustice, and R.~Vasudevan, ``Pairwise consistent measurement set maximization for robust multi-robot map merging,'' in \emph{Proceedings of the IEEE International Conference on Robotics and Automation (ICRA)}, 2018, pp. 2916--2923.

\bibitem[Shan et~al.(2020)Shan, Englot, Meyers, Wang, Ratti, and Daniela]{slam_liosam2020shan}
T.~Shan, B.~Englot, D.~Meyers, W.~Wang, C.~Ratti, and R.~Daniela, ``{LIO-SAM}: Tightly-coupled lidar inertial odometry via smoothing and mapping,'' in \emph{Proceedings of the IEEE/RSJ International Conference on Intelligent Robots and Systems (IROS)}, 2020, pp. 5135--5142.

\bibitem[Gu et~al.(2024)Gu, Kuwajerwala, Morin, Jatavallabhula, Sen, Agarwal, Rivera, Paul, Ellis, Chellappa, Gan, De~Melo, Tenenbaum, Torralba, Shkurti, and Paull]{guConceptGraphsOpenVocabulary2024}
Q.~Gu, A.~Kuwajerwala, S.~Morin, K.~M. Jatavallabhula, B.~Sen, A.~Agarwal, C.~Rivera, W.~Paul, K.~Ellis, R.~Chellappa, C.~Gan, C.~M. De~Melo, J.~B. Tenenbaum, A.~Torralba, F.~Shkurti, and L.~Paull, ``{ConceptGraphs}: Open-vocabulary {3D} scene graphs for perception and planning,'' in \emph{Proceedings of the IEEE International Conference on Robotics and Automation (ICRA)}, 2024, pp. 5021--5028.

\bibitem[Werby et~al.(2024)Werby, Huang, Büchner, Valada, and Burgard]{werby23hovsg}
A.~Werby, C.~Huang, M.~Büchner, A.~Valada, and W.~Burgard, ``Hierarchical open-vocabulary {3D} scene graphs for language-grounded robot navigation,'' \emph{Robotics: Science and Systems}, 2024.

\bibitem[Qiu et~al.(2024)Qiu, Song, Peng, Suryadevara, Yang, Liu, Ji, Jia, Yang, Zou, and Wang]{qiu-song-peng-2024-wildlma}
\BIBentryALTinterwordspacing
R.-Z. Qiu, Y.~Song, X.~Peng, S.~A. Suryadevara, G.~Yang, M.~Liu, M.~Ji, C.~Jia, R.~Yang, X.~Zou, and X.~Wang, ``{WildLMa}: Long horizon loco-manipulation in the wild,'' 2024. [Online]. Available: \url{https://arxiv.org/abs/2411.15131}
\BIBentrySTDinterwordspacing

\bibitem[Devarakonda et~al.(2024)Devarakonda, Goswami, Kaypak, Patel, Khorrambakht, Krishnamurthy, and Khorrami]{Devarakonda2024OrionNavOP}
\BIBentryALTinterwordspacing
V.~N. Devarakonda, R.~G. Goswami, A.~U. Kaypak, N.~Patel, R.~Khorrambakht, P.~Krishnamurthy, and F.~Khorrami, ``{OrionNav}: Online planning for robot autonomy with context-aware {LLM} and open-vocabulary semantic scene graphs,'' 2024. [Online]. Available: \url{https://arxiv.org/abs/2410.06239}
\BIBentrySTDinterwordspacing

\bibitem[Steinke et~al.(2025)Steinke, Büchner, Vödisch, and Valada]{steinkeCollaborativeDynamic2025b}
\BIBentryALTinterwordspacing
T.~Steinke, M.~Büchner, N.~Vödisch, and A.~Valada, ``Collaborative dynamic {3D} scene graphs for open-vocabulary urban scene understanding,'' 2025. [Online]. Available: \url{https://arxiv.org/abs/2503.08474}
\BIBentrySTDinterwordspacing

\bibitem[Armeni et~al.(2019)Armeni, He, Zamir, Gwak, Malik, Fischer, and Savarese]{armeni3DScene2019b}
I.~Armeni, Z.-Y. He, A.~Zamir, J.~Gwak, J.~Malik, M.~Fischer, and S.~Savarese, ``{3D} scene graph: A structure for unified semantics, {3D} space, and camera,'' in \emph{Proceedings of the IEEE/CVF International Conference on Computer Vision (ICCV)}, 2019, pp. 5663--5672.

\bibitem[Rosinol et~al.(2020)Rosinol, Abate, Chang, and Carlone]{rosinolKimera2020}
A.~Rosinol, M.~Abate, Y.~Chang, and L.~Carlone, ``Kimera: an open-source library for real-time metric-semantic localization and mapping,'' in \emph{Proceedings of the IEEE International Conference on Robotics and Automation (ICRA)}, 2020, pp. 1689--1696.

\bibitem[Rosinol et~al.()Rosinol, Gupta, Abate, Shi, and Carlone]{rosinol3DDynamic2020}
\BIBentryALTinterwordspacing
A.~Rosinol, A.~Gupta, M.~Abate, J.~Shi, and L.~Carlone, ``{3D} dynamic scene graphs: Actionable spatial perception with places, objects, and humans.'' [Online]. Available: \url{https://arxiv.org/abs/2002.06289}
\BIBentrySTDinterwordspacing

\bibitem[Hughes et~al.(2022)Hughes, Chang, and Carlone]{hughes2022hydra}
N.~Hughes, Y.~Chang, and L.~Carlone, ``Hydra: A real-time spatial perception system for {3D} scene graph construction and optimization,'' \emph{Robotics: Science and Systems (RSS)}, 2022.

\bibitem[Bavle et~al.(2022)Bavle, Sanchez-Lopez, Shaheer, Civera, and Voos]{bavle2022sgraphs+}
\BIBentryALTinterwordspacing
H.~Bavle, J.~L. Sanchez-Lopez, M.~Shaheer, J.~Civera, and H.~Voos, ``{S-Graphs+}: Real-time localization and mapping leveraging hierarchical representations,'' 2022. [Online]. Available: \url{https://arxiv.org/abs/2212.11770}
\BIBentrySTDinterwordspacing

\bibitem[Bavle et~al.(2025)Bavle, Sanchez-Lopez, Shaheer, Civera, and Voos]{bavle2025sgraphs20hierarchicalsemantic}
\BIBentryALTinterwordspacing
------, ``{S-Graphs} 2.0 -- a hierarchical-semantic optimization and loop closure for {SLAM},'' 2025. [Online]. Available: \url{https://arxiv.org/abs/2502.18044}
\BIBentrySTDinterwordspacing

\bibitem[Takmaz et~al.(2024)Takmaz, Delitzas, Sumner, Engelmann, Wald, and Tombari]{takmazSearch3DHierarchical2024}
\BIBentryALTinterwordspacing
A.~Takmaz, A.~Delitzas, R.~W. Sumner, F.~Engelmann, J.~Wald, and F.~Tombari, ``{Search3D}: Hierarchical open-vocabulary {3D} segmentation,'' 2024. [Online]. Available: \url{https://arxiv.org/abs/2409.18431}
\BIBentrySTDinterwordspacing

\bibitem[Xu et~al.(2024)Xu, Luo, Wang, Kamat, and Menassa]{xu2024point2graphendtoendpointcloudbased}
\BIBentryALTinterwordspacing
Y.~Xu, Z.~Luo, Q.~Wang, V.~Kamat, and C.~Menassa, ``{Point2Graph}: An end-to-end point cloud-based {3D} open-vocabulary scene graph for robot navigation,'' 2024. [Online]. Available: \url{https://arxiv.org/abs/2409.10350}
\BIBentrySTDinterwordspacing

\bibitem[Maggio et~al.(2024)Maggio, Chang, Hughes, Trang, Griffith, Dougherty, Cristofalo, Schmid, and Carlone]{maggio2024Clio}
D.~Maggio, Y.~Chang, N.~Hughes, M.~Trang, D.~Griffith, C.~Dougherty, E.~Cristofalo, L.~Schmid, and L.~Carlone, ``Clio: Real-time task-driven open-set {3D} scene graphs,'' \emph{IEEE Robotics and Automation Letters}, vol.~9, no.~10, pp. 8921--8928, 2024.

\bibitem[Gao et~al.(2024)Gao, Tang, Wang, Li, and Yu]{gaoEnhancingScene2024}
F.~Gao, J.~Tang, J.~Wang, S.~Li, and J.~Yu, ``Enhancing scene understanding for vision-and-language navigation by knowledge awareness,'' \emph{IEEE Robotics and Automation Letters}, vol.~9, no.~12, pp. 10\,874--10\,881, 2024.

\bibitem[Xu et~al.(2025)Xu, Ila, Zhou, and Jin]{xu2025tbhsu}
\BIBentryALTinterwordspacing
W.~Xu, V.~Ila, L.~Zhou, and C.~T. Jin, ``{TB-HSU}: Hierarchical {3D} scene understanding with contextual affordances,'' 2025. [Online]. Available: \url{https://arxiv.org/abs/2412.05596}
\BIBentrySTDinterwordspacing

\bibitem[Kassab et~al.(2024)Kassab, Mattamala, Morin, Büchner, Valada, Paull, and Fallon]{kassab2024barenecessities}
\BIBentryALTinterwordspacing
C.~Kassab, M.~Mattamala, S.~Morin, M.~Büchner, A.~Valada, L.~Paull, and M.~Fallon, ``The bare necessities: Designing simple, effective open-vocabulary scene graphs,'' 2024. [Online]. Available: \url{https://arxiv.org/abs/2412.01539}
\BIBentrySTDinterwordspacing

\bibitem[Tang et~al.(2025)Tang, Wang, Deng, Zheng, Deng, and Yue]{tangOpenINOpenVocabulary2025}
\BIBentryALTinterwordspacing
Y.~Tang, M.~Wang, Y.~Deng, Z.~Zheng, J.~Deng, and Y.~Yue, ``{{OpenIN}}: Open-vocabulary instance-oriented navigation in dynamic domestic environments,'' 2025. [Online]. Available: \url{https://arxiv.org/abs/2501.04279}
\BIBentrySTDinterwordspacing

\bibitem[202(2025)]{2025topofield}
``{TOPO-FIELD}: Topometric mapping with braininspired hierarchical layout-object-position fields,'' in \emph{Proceedings of the International Conference on Learning Representations (ICLR)}, 2025.

\bibitem[Berg et~al.(2022)Berg, Konidaris, and Tellex]{bergUsingLanguage2022}
M.~Berg, G.~Konidaris, and S.~Tellex, ``Using language to generate state abstractions for long-range planning in outdoor environments,'' in \emph{Proceedings of the International Conference on Robotics and Automation (ICRA)}, 2022, pp. 1888--1895.

\bibitem[Greve et~al.(2024)Greve, Büchner, Vödisch, Burgard, and Valada]{greve2024curb}
E.~Greve, M.~Büchner, N.~Vödisch, W.~Burgard, and A.~Valada, ``Collaborative dynamic {3D} scene graphs for automated driving,'' in \emph{Proceedings of the IEEE International Conference on Robotics and Automation (ICRA)}, 2024, pp. 11\,118--11\,124.

\bibitem[Strader et~al.(2024)Strader, Hughes, Chen, Speranzon, and Carlone]{straderIndoorOutdoor2024}
J.~Strader, N.~Hughes, W.~Chen, A.~Speranzon, and L.~Carlone, ``Indoor and outdoor {3D} scene graph generation via language-enabled spatial ontologies,'' \emph{IEEE Robotics and Automation Letters}, vol.~9, no.~6, pp. 4886--4893, 2024.

\bibitem[Ray et~al.(2024)Ray, Bradley, Carlone, and Roy]{rayTaskMotion2024}
\BIBentryALTinterwordspacing
A.~Ray, C.~Bradley, L.~Carlone, and N.~Roy, ``Task and motion planning in hierarchical {3D} scene graphs,'' 2024. [Online]. Available: \url{https://arxiv.org/abs/2403.08094}
\BIBentrySTDinterwordspacing

\bibitem[Krishna et~al.(2017)Krishna, Zhu, Groth, Johnson, Hata, Kravitz, Chen, Kalantidis, Li, Shamma, Bernstein, and Fei{-}Fei]{krishna2DSG2017}
R.~Krishna, Y.~Zhu, O.~Groth, J.~Johnson, K.~Hata, J.~Kravitz, S.~Chen, Y.~Kalantidis, L.~Li, D.~A. Shamma, M.~S. Bernstein, and L.~Fei{-}Fei, ``Visual genome: Connecting language and vision using crowdsourced dense image annotations,'' \emph{International Journal of Computer Vision}, vol. 123, no.~1, pp. 32--73, 2017.

\bibitem[Armeni et~al.(2016)Armeni, Sener, Zamir, Jiang, Brilakis, Fischer, and Savarese]{armeni3DSemantic2016}
I.~Armeni, O.~Sener, A.~R. Zamir, H.~Jiang, I.~Brilakis, M.~Fischer, and S.~Savarese, ``{3D} semantic parsing of large-scale indoor spaces,'' in \emph{Proceedings of the IEEE Conference on Computer Vision and Pattern Recognition (CVPR)}, 2016, pp. 1534--1543.

\bibitem[OpenAI(2024)]{OpenAI2024GPT4o}
\BIBentryALTinterwordspacing
OpenAI, ``{GPT-4o} technical report,'' 2024. [Online]. Available: \url{https://platform.openai.com/docs/models/gpt-4o}
\BIBentrySTDinterwordspacing

\bibitem[Touvron et~al.(2023)Touvron, Lavril, Izacard, Martinet, Lachaux, Lacroix, Rozière, Goyal, Hambro, Azhar, Rodriguez, Joulin, Grave, and Lample]{touvron2023llamaopenefficientfoundation}
\BIBentryALTinterwordspacing
H.~Touvron, T.~Lavril, G.~Izacard, X.~Martinet, M.-A. Lachaux, T.~Lacroix, B.~Rozière, N.~Goyal, E.~Hambro, F.~Azhar, A.~Rodriguez, A.~Joulin, E.~Grave, and G.~Lample, ``{LLaMA}: Open and efficient foundation language models,'' 2023. [Online]. Available: \url{https://arxiv.org/abs/2302.13971}
\BIBentrySTDinterwordspacing

\bibitem[Radford et~al.(2021)Radford, Kim, Hallacy, Ramesh, Goh, Agarwal, Sastry, Askell, Mishkin, Clark, Krueger, and Sutskever]{radfordLearningTransferable2021}
A.~Radford, J.~W. Kim, C.~Hallacy, A.~Ramesh, G.~Goh, S.~Agarwal, G.~Sastry, A.~Askell, P.~Mishkin, J.~Clark, G.~Krueger, and I.~Sutskever, ``Learning transferable visual models from natural language supervision,'' in \emph{Proceedings of the International Conference on Machine Learning (ICML)}, 2021, pp. 8748--8763.

\bibitem[Jatavallabhula et~al.(2023)Jatavallabhula, Kuwajerwala, Gu, Omama, Chen, Li, Iyer, Saryazdi, Keetha, Tewari, Tenenbaum, {de Melo}, Krishna, Paull, Shkurti, and Torralba]{jatavallabhula2023Conceptfusion}
K.~Jatavallabhula, A.~Kuwajerwala, Q.~Gu, M.~Omama, T.~Chen, S.~Li, G.~Iyer, S.~Saryazdi, N.~Keetha, A.~Tewari, J.~Tenenbaum, C.~{de Melo}, M.~Krishna, L.~Paull, F.~Shkurti, and A.~Torralba, ``{ConceptFusion}: Open-set multimodal {3D} mapping,'' \emph{Robotics: Science and Systems (RSS)}, 2023.

\bibitem[Kirillov et~al.(2023)Kirillov, Mintun, Ravi, Mao, Rolland, Gustafson, Xiao, Whitehead, Berg, Lo, Dollár, and Girshick]{kirillov2023sam}
A.~Kirillov, E.~Mintun, N.~Ravi, H.~Mao, C.~Rolland, L.~Gustafson, T.~Xiao, S.~Whitehead, A.~C. Berg, W.-Y. Lo, P.~Dollár, and R.~Girshick, ``Segment anything,'' in \emph{Proceedings of the IEEE/CVF International Conference on Computer Vision (ICCV)}, 2023, pp. 3992--4003.

\bibitem[Ravi et~al.(2024)Ravi, Gabeur, Hu, Hu, Ryali, Ma, Khedr, R{\"a}dle, Rolland, Gustafson, Mintun, Pan, Alwala, Carion, Wu, Girshick, Doll{\'a}r, and Feichtenhofer]{ravi2024sam2}
\BIBentryALTinterwordspacing
N.~Ravi, V.~Gabeur, Y.-T. Hu, R.~Hu, C.~Ryali, T.~Ma, H.~Khedr, R.~R{\"a}dle, C.~Rolland, L.~Gustafson, E.~Mintun, J.~Pan, K.~V. Alwala, N.~Carion, C.-Y. Wu, R.~Girshick, P.~Doll{\'a}r, and C.~Feichtenhofer, ``{SAM 2}: Segment anything in images and videos,'' 2024. [Online]. Available: \url{https://arxiv.org/abs/2408.00714}
\BIBentrySTDinterwordspacing

\bibitem[Zhao et~al.(2023)Zhao, Ding, An, Du, Yu, Li, Tang, and Wang]{zhao2023fastsam}
\BIBentryALTinterwordspacing
X.~Zhao, W.~Ding, Y.~An, Y.~Du, T.~Yu, M.~Li, M.~Tang, and J.~Wang, ``Fast segment anything,'' 2023. [Online]. Available: \url{https://arxiv.org/abs/2306.12156}
\BIBentrySTDinterwordspacing

\bibitem[Zhang et~al.(2023)Zhang, Han, Qiao, Kim, Bae, Lee, and Hong]{zhang2023mobilesam}
\BIBentryALTinterwordspacing
C.~Zhang, D.~Han, Y.~Qiao, J.~U. Kim, S.-H. Bae, S.~Lee, and C.~S. Hong, ``Faster segment anything: Towards lightweight {SAM} for mobile applications,'' 2023. [Online]. Available: \url{https://arxiv.org/abs/2306.14289}
\BIBentrySTDinterwordspacing

\bibitem[Cheng et~al.(2024)Cheng, Song, Ge, Liu, Wang, and Shan]{cheng2024YOLOWorld}
T.~Cheng, L.~Song, Y.~Ge, W.~Liu, X.~Wang, and Y.~Shan, ``{YOLO-World}: Real-time open-vocabulary object detection,'' in \emph{Proceedings of the IEEE Conference on Computer Vision and Pattern Recognition (CVPR)}, 2024.

\bibitem[Valada et~al.()Valada, Oliveira, Brox, and Burgard]{valadaFreiburgForest2017a}
A.~Valada, G.~L. Oliveira, T.~Brox, and W.~Burgard, ``Deep multispectral semantic scene understanding of forested environments using multimodal fusion,'' in \emph{2016 International Symposium on Experimental Robotics}, D.~Kulić, Y.~Nakamura, O.~Khatib, and G.~Venture, Eds., vol.~1, pp. 465--477.

\bibitem[Maturana et~al.()Maturana, Chou, Uenoyama, and Scherer]{maturanaRealTimeSemantic2018a}
D.~Maturana, P.-W. Chou, M.~Uenoyama, and S.~Scherer, ``Real-time semantic mapping for autonomous off-road navigation,'' in \emph{Field and {{Service Robotics}}}, M.~Hutter and R.~Siegwart, Eds.\hskip 1em plus 0.5em minus 0.4em\relax Springer International Publishing, vol.~5, pp. 335--350.

\bibitem[Wigness et~al.(2019)Wigness, Eum, Rogers, Han, and Kwon]{wignessRUGDDataset2019}
M.~Wigness, S.~Eum, J.~G. Rogers, D.~Han, and H.~Kwon, ``A {RUGD} dataset for autonomous navigation and visual perception in unstructured outdoor environments,'' in \emph{Proceedings of the {IEEE} Conference on Intelligent Robots and Systems (IROS)}, 2019, pp. 5000--5007.

\bibitem[Jiang et~al.(2021)Jiang, Osteen, Wigness, and Saripalli]{jiangRELLIS3DDataset2021}
P.~Jiang, P.~Osteen, M.~Wigness, and S.~Saripalli, ``{RELLIS-3D} dataset: Data, benchmarks and analysis,'' in \emph{Proceedings of the IEEE International Conference on Robotics and Automation (ICRA)}, 2021, pp. 1110--1116.

\bibitem[Metzger et~al.(2021)Metzger, Mortimer, and Wuensche]{metzger2021}
K.~A. Metzger, P.~Mortimer, and H.-J. Wuensche, ``A fine-grained dataset and its efficient semantic segmentation for unstructured driving scenarios,'' in \emph{Proceedings of the International Conference on Pattern Recognition (ICPR)}, 2021, pp. 7892--7899.

\bibitem[Mortimer et~al.(2024)Mortimer, Hagmanns, Granero, Luettel, Petereit, and Wuensche]{mortimer2024goose}
P.~Mortimer, R.~Hagmanns, M.~Granero, T.~Luettel, J.~Petereit, and H.-J. Wuensche, ``The {GOOSE} dataset for perception in unstructured environments,'' in \emph{Proceedings of the IEEE International Conference on Robotics and Automation (ICRA)}, 2024.

\bibitem[Hagmanns et~al.(2024)Hagmanns, Mortimer, Granero, Luettel, and Petereit]{hagmanns2024gooseEx}
\BIBentryALTinterwordspacing
R.~Hagmanns, P.~Mortimer, M.~Granero, T.~Luettel, and J.~Petereit, ``Excavating in the wild: The {GOOSE-Ex} dataset for semantic segmentation,'' 2024. [Online]. Available: \url{https://arxiv.org/abs/2409.18788}
\BIBentrySTDinterwordspacing

\bibitem[Vidanapathirana et~al.(2024)Vidanapathirana, Knights, Hausler, Cox, Ramezani, Jooste, Griffiths, Mohamed, Sridharan, Fookes, and Moghadam]{vidanapathiranaWildScenesbenchmark2024}
K.~Vidanapathirana, J.~Knights, S.~Hausler, M.~Cox, M.~Ramezani, J.~Jooste, E.~Griffiths, S.~Mohamed, S.~Sridharan, C.~Fookes, and P.~Moghadam, ``{WildScenes}: A benchmark for {2D} and {3D} semantic segmentation in large-scale natural environments,'' \emph{The International Journal of Robotics Research}, 2024.

\bibitem[Zhang and Singh(2014)]{slam_loam2014zhang}
J.~Zhang and S.~Singh, ``{LOAM}: Lidar odometry and mapping in real-time,'' \emph{In Robotics: Science and Systems}, 2014.

\bibitem[Varghese and M.(2024)]{yolov8_ultralytics}
R.~Varghese and S.~M., ``{YOLOv8}: A novel object detection algorithm with enhanced performance and robustness,'' in \emph{Proceedings of the International Conference on Advances in Data Engineering and Intelligent Computing Systems (ADICS)}, 2024, pp. 1--6.

\bibitem[Bentley(1975)]{bentley1975kdtrees}
J.~L. Bentley, ``Multidimensional binary search trees used for associative searching,'' \emph{Communications of the ACM}, vol.~18, no.~9, p. 509–517, 9 1975.

\bibitem[Lau et~al.(2013)Lau, Sprunk, and Burgard]{lauEfficientgridbased2013}
B.~Lau, C.~Sprunk, and W.~Burgard, ``Efficient grid-based spatial representations for robot navigation in dynamic environments,'' \emph{Robotics and Autonomous Systems}, vol.~61, no.~10, pp. 1116--1130, 2013.

\end{thebibliography}
\end{document}